\documentclass[runningheads]{llncs}
\usepackage[T1]{fontenc}
\usepackage{graphicx}
\usepackage{booktabs}
\usepackage[misc]{ifsym}
\newcommand{\corr}{(\Letter)}

\usepackage{algorithm}
\usepackage{tcolorbox}
\usepackage{xcolor}
\usepackage{multirow}
\usepackage{tabularx}
\usepackage{amsmath}
\usepackage{booktabs}
\usepackage{pifont}
\usepackage{graphicx} 
\usepackage{subcaption} 
\usepackage{hyperref}

\usepackage{fontawesome}
\usepackage{url}
\usepackage{enumitem} 

\begin{document}

\title{Iterative Hypothesis Generation for Scientific Discovery with Monte Carlo Nash Equilibrium Self-Refining Trees}

\titlerunning{Iterative Hypothesis Generation for Scientific Discovery with MC-NEST}

\author{
Gollam Rabby\inst{1 \dagger} \corr \and
Diyana Muhammed \inst{2 \dagger} \and
Prasenjit Mitra\inst{2} \and
Sören Auer\inst{1,2}} 

\authorrunning{G. Rabby et al.}

\institute{L3S Research Center, Leibniz University Hannover, Hannover, Germany \email{\{gollam.rabby, mitra\}@l3s.de}
\and
TIB—Leibniz Information Centre for Science and Technology, Hannover, Germany\email{\{diyana.muhammed, auer\}@tib.eu}
\newline
 \vspace{0.2cm}
\raisebox{-1pt}{\faDatabase} \href{https://huggingface.co/datasets/tourist800/LLM4CSHypoGen}{\texttt{Dataset}} \quad 
\raisebox{-1pt}{\faGlobe} \href{https://corei5.github.io/Hypothesis_Generation_With_MC_NEST/}{\texttt{Project Page}} \quad 
\raisebox{-1pt}{\faGithub} \href{https://github.com/corei5/Hypothesis_Generation_With_MC_NEST}{\texttt{Codebase}}
}

\footnotetext{\hspace{-0.4cm}$^\dagger$ Equal contribution.}


\maketitle              

\vspace{-6mm}
\begin{abstract}

Scientific hypothesis generation is a fundamentally challenging task in research, requiring the synthesis of novel and empirically grounded insights. Traditional approaches rely on human intuition and domain expertise, while purely large language model (LLM) based methods often struggle to produce hypotheses that are both innovative and reliable. To address these limitations, we propose the Monte Carlo Nash Equilibrium Self-Refine Tree (MC-NEST), a novel framework that integrates Monte Carlo Tree Search (MCTS) with Nash Equilibrium strategies to iteratively refine and validate hypotheses. MC-NEST dynamically balances exploration and exploitation through adaptive sampling strategies, which prioritize high-potential hypotheses while maintaining diversity in the search space. We demonstrate the effectiveness of MC-NEST through comprehensive experiments across multiple domains, including biomedicine, social science, and computer science. MC-NEST achieves average scores of 2.65, 2.74, and 2.80 (on a 1-3 scale) for novelty, clarity, significance, and verifiability metrics on the social science, computer science, and biomedicine datasets, respectively, outperforming state-of-the-art prompt-based methods, which achieve 2.36, 2.51, and 2.52 on the same datasets. These results underscore MC-NEST's ability to generate high-quality, empirically grounded hypotheses across diverse domains. Furthermore, MC-NEST facilitates structured human-AI collaboration, ensuring that LLMs augment human creativity rather than replace it. By addressing key challenges such as iterative refinement and the exploration-exploitation balance, MC-NEST sets a new benchmark in automated hypothesis generation. The framework provides a robust and adaptable approach that advances the boundaries of scientific discovery. Additionally, MC-NEST's ethical design enables responsible AI use, emphasizing transparency and human supervision in hypothesis generation.

\end{abstract}


\section{Introduction}
\label{sec:Introduction}

Scientific hypothesis generation drives discovery and innovation but remains limited by the scale and complexity of modern challenges. While large language models (LLMs) show promise in automating this process~\cite{brown2020language}, existing approaches struggle to generate hypotheses that are both novel and empirically grounded due to a lack of iterative refinement and poor exploration-exploitation balance~\cite{flaspohler2022balancing}. To address these challenges, we utilize the Monte Carlo Nash Equilibrium Self-Refine Tree (MC-NEST), a framework that integrates the Monte Carlo Tree Search (MCTS) with Nash Equilibrium strategies to iteratively refine hypotheses~\cite{DBLP:journals/corr/abs-2411-15645}. MC-NEST frames hypothesis generation as a game, where the players are competing strategies for exploring and refining hypotheses. This game-theoretic approach allows MC-NEST to balance the trade-offs between exploring new ideas and exploiting known high-quality hypotheses. Each strategy aims to maximize the quality of the generated hypotheses, and Nash Equilibrium ensures a balance where no player (strategy) can improve its outcome by unilaterally changing its approach. These strategies guide the exploration and refinement phases by dynamically adjusting the trade-off between exploring new hypotheses and exploiting known high-quality ones, ensuring optimal hypothesis generation. 

MC-NEST dynamically balances exploration and exploitation using adaptive sampling techniques, ensuring diverse and high-potential hypotheses. The framework operates in two phases: (1) \textit{an exploration phase, where MCTS navigates the hypothesis space guided by Nash Equilibrium,} and (2) \textit{a refinement phase, where adaptive sampling and iterative self-reflection ensure hypotheses are innovative and empirically grounded.} For instance, in peptide optimization, exploration might involve proposing a new substitution (e.g., replacing arginine with lysine) to test its effect on solubility, while exploitation would refine this idea by validating whether the substitution improves solubility without compromising the peptide's nuclear localization function. Experiments across biomedicine, social science, and computer science demonstrate MC-NEST's effectiveness in hypothesis generation. MC-NEST achieves higher novelty, clarity, significance, and verifiability compared to existing methods~\cite{lee2022scibench}, demonstrating its effectiveness in generating scientifically impactful hypotheses. Specifically, MC-NEST achieves scores of 2.65, 2.74, and 2.80 (on a 1-3 scale) for novelty, clarity, significance, and verifiability on the social science, computer science, and biomedicine datasets, respectively. These results outperform state-of-the-art prompt-based methods, which achieve 2.36, 2.51, and 2.52 on the same datasets. This improvement demonstrates MC-NEST's ability to generate hypotheses that are not only innovative but also empirically grounded and scientifically impactful. 

A key innovation is MC-NEST's ability to incorporate emerging scientific literature, addressing the limitations of automatic refinement and exploration-exploitation balance. The framework supports structured human-AI collaboration, where LLMs augment human expertise rather than replace it. This approach balances AI and human judgment, mitigating over-reliance on AI. While AI excels at generating novel hypotheses and exploring chemical spaces, human expertise is critical for interpreting results, identifying biases, and ensuring ethical decisions. For example, in peptide optimization, MC-NEST proposes substitutions (e.g., lysine-for-arginine) to improve solubility, while humans validate whether these changes maintain nuclear localization and align with biochemical principles. This iterative collaboration combines AI's exploratory capabilities with human expertise, ensuring scientifically robust and ethically sound outcomes. 

\begin{figure}[htbp]
    \centering
    \begin{subfigure}[b]{0.4\textwidth}
        \centering
        \includegraphics[width=\textwidth]{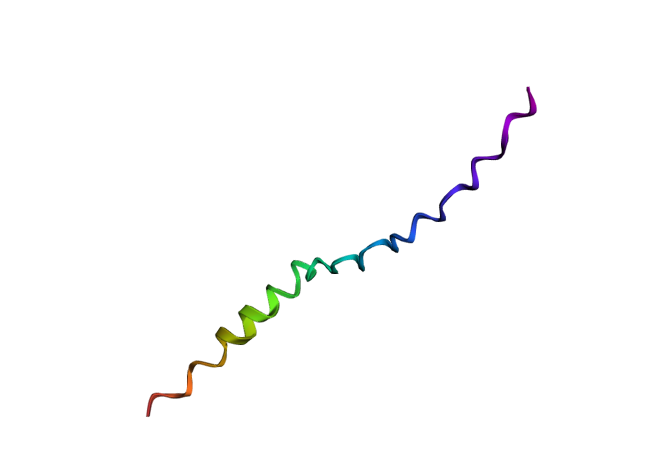}
        \caption{Original peptide} 
        \label{fig:original_peptide}
    \end{subfigure}
    \hfill
    \begin{subfigure}[b]{0.4\textwidth}
        \centering
        \includegraphics[width=\textwidth]{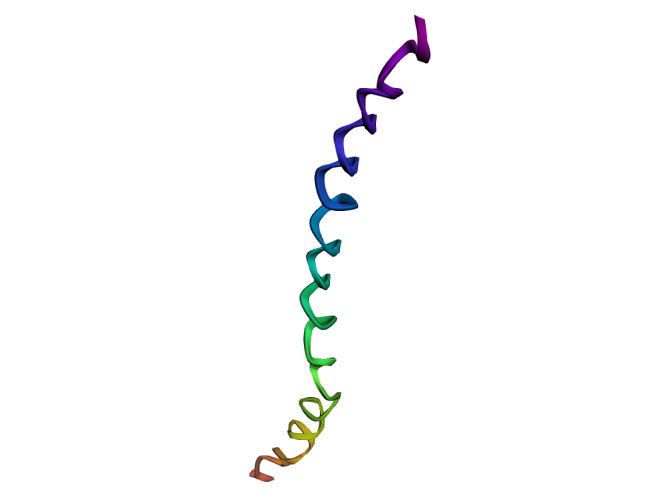}
        \caption{MC-NEST updated peptide} 
        \label{fig:updated_peptide}
    \end{subfigure}
    \caption{
        Comparison of original and MC-NEST hypothesis-generated synthetic peptide sequences visualized by AlphaFold~\cite{jumper2021highly}. 
        (a) Original sequence with NLS (red) and glycine-rich linker (blue). 
        (b) Updated sequence with lysine-for-arginine substitutions in the NLS (red) and alanine-for-glycine substitutions in the linker (blue). 
        Code: \href{https://colab.research.google.com/drive/1B7P0ZVaLhGLRF2EkK4BsdpadpbG_dJkx?usp=sharing}{Google Colab Notebook}
        }
    \label{fig:peptide_comparison}
\end{figure}

For research problems, impact is as critical as novelty. While novelty ensures that hypotheses are original, impact ensures they address meaningful scientific challenges. MC-NEST achieves this balance by generating hypotheses that are not only novel but also grounded in domain-specific knowledge and validated for real-world applicability. Unlike purely exploratory methods, MC-NEST incorporates iterative refinement and validation, ensuring that hypotheses are both innovative and empirically grounded. For example, in complex scientific domains such as protein engineering, MC-NEST's proposed modifications (e.g., lysine-for-arginine substitutions) are designed to enhance solubility while maintaining critical functional properties—a dual focus that directly addresses high-priority scientific and therapeutic needs. By combining exploration with rigorous validation, MC-NEST ensures that its hypotheses are not only novel but also impactful, contributing to solving real-world problems with significant scientific and practical implications. Our contributions include:
\begin{itemize}
    \item MC-NEST, a framework integrating MCTS and Nash Equilibrium for hypothesis generation, enhanced by adaptive sampling techniques.
    \item A comprehensive performance analysis across multiple domains, with detailed studies highlighting the impact of each component.
    \item A human-AI collaboration approach that improves hypothesis quality through expert refinement.
\end{itemize}

To ensure reproducibility, we will release all used source codes, datasets, and evaluation protocols.

To illustrate MC-NEST's capabilities, we present an example of hypothesis generation and refinement for optimizing a synthetic peptide sequence (\textit{MARTKQTARKSTGGKAPRKQLASKAARKSAARAAAAGGGGGGG}) for nuclear localization and solubility. MC-NEST generates an initial hypothesis: \textit{Substituting lysine for arginine in the nuclear localization signal (NLS) preserves the positive charge required for nuclear import while enhancing solubility due to lysine's less bulky structure.} Validation against biochemical principles reveals potential trade-offs, such as reduced binding affinity to nuclear import receptors~\cite{hahn2011importin}. MC-NEST refines the hypothesis by incorporating additional modifications: \textit{Replacing some glycine residues with alanine in the glycine-rich linker to maintain flexibility without introducing phosphorylation sites.} Experimental validation confirms that the modified peptide outperforms the original sequence, retaining nuclear localization efficiency while improving solubility and functionality. The updated sequence generated by MC-NEST is: \textit{MAKTQTGRPKSTGGPAPRKQLASPPARKSVAARAAAASGGGSGG}. A visual comparison (by AlphaFold) of the original and updated peptide sequences is shown in Figure~\ref{fig:peptide_comparison}.

\begin{figure*}[ht]
    \centering
    \includegraphics[height=4cm, width=\textwidth]{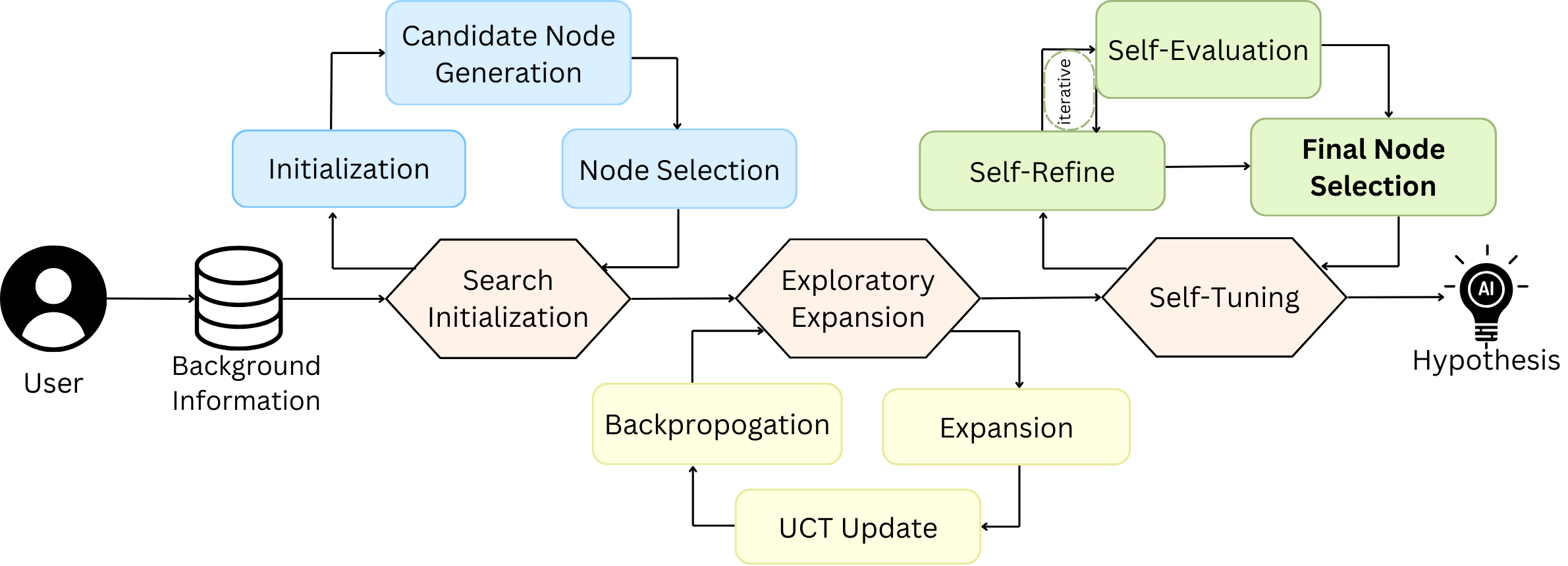}
    \caption{Overview of the MC-NEST methodology for hypothesis generation.}
    \label{fig:methodology_overview}
\end{figure*}

\section{Methodology}
\label{sub:what_MC_NEST}

MC-NEST is a computational framework designed to enhance the problem-solving capabilities of LLMs for scientific hypothesis generation~\cite{DBLP:journals/corr/abs-2411-15645}. As illustrated in Figure~\ref{fig:methodology_overview}, MC-NEST integrates the Monte Carlo Tree Search, a decision-making algorithm for exploring large search spaces~\cite{browne2012survey}, with Nash Equilibrium strategies to iteratively refine hypotheses and solutions. By dynamically balancing exploration and exploitation, MC-NEST ensures that generated hypotheses are both innovative and empirically grounded. 

\subsubsection{Problem Setting for Hypothesis Generation.}
MC-NEST is designed for a structured search over combinatorial hypothesis spaces, particularly in domains requiring rigorous reasoning and insight. The framework addresses the challenge of efficiently navigating vast search spaces while ensuring quality, efficiency, and novelty. Specifically, MC-NEST targets problems where:

\begin{itemize}
    \item \textit{The hypothesis space is combinatorial}, with solutions constructed from smaller reasoning steps or building blocks. For example, in protein engineering, a hypothesis might propose amino acid substitutions to optimize functions like nuclear localization or solubility~\cite{wang2023scientific}. A specific hypothesis could suggest substituting lysine for arginine in a nuclear localization signal (NLS), preserving the positive charge required for nuclear import while enhancing solubility due to lysine's less bulky structure. Such hypotheses are built from testable steps (e.g., charge preservation, solubility enhancement) that can be experimentally validated.

    \item \textit{The search space is large for exhaustive exploration}, necessitating intelligent traversal strategies~\cite{kocsis2006bandit}. For example, the space of possible amino acid substitutions is intractable without a guided search. Traditional methods often focus on well-known substitutions (e.g., arginine-to-lysine in NLS), while MC-NEST explores less-studied modifications, such as introducing alanine into glycine-rich linkers to enhance flexibility without adding phosphorylation sites. By prioritizing high-potential but underexplored changes, MC-NEST uncovers novel solutions missed by traditional approaches.

    \item \textit{Solutions must satisfy strict correctness criteria}, including clarity, testability, relevance, and novelty~\cite{lee2022scibench}. For instance, a hypothesis must clearly describe relationships (e.g., ``substituting lysine for arginine enhances nuclear import efficiency''), be testable (e.g., via fluorescence microscopy or solubility assays), relevant (e.g., optimizing synthetic peptides for mammalian cell expression), and novel (e.g., identifying alanine's role in linker flexibility). MC-NEST ensures that hypotheses meet these criteria by iteratively refining and validating them against biochemical principles and experimental data.
\end{itemize}

\subsubsection{Search Space and Traversal Strategy.}
\label{subsec:Search_Space}


The search space in MC-NEST is represented as a tree, where nodes correspond to solutions (e.g., hypotheses or amino acid substitutions), and edges represent logical transitions. The traversal strategy combines exploration and exploitation: 1) \textit{Upper Confidence Bound for Trees (UCT)} balances exploration and exploitation by estimating branch potential using confidence intervals, favoring high-uncertainty or high-performance paths~\cite{DBLP:journals/corr/abs-2411-15645} (\autoref{subsec:MC_NEST}). For example, UCT explores less-studied substitutions (e.g., alanine in glycine-rich linkers) while leveraging known modifications (e.g., lysine-for-arginine in the NLS). 2) \textit{Exploration} prioritizes underexplored branches, balancing novelty and promise, as seen in game-playing AI like AlphaGo~\cite{silver2016mastering} (\autoref{subsec:MC_NEST}). 3) \textit{Exploitation} refines promising branches using probabilistic node selection, focusing on high-quality regions while maintaining diversity (\autoref{subsec:MC_NEST}). For instance, MC-NEST exploits beneficial substitutions (e.g., lysine-for-arginine) while exploring novel combinations (e.g., alanine in glycine-rich linkers) to optimize functionality.

\subsection{Benefits of the MC-NEST Framework in Hypothesis Generation}
\label{sec:Benefits_of_mc_nest}

Scientific discovery has traditionally relied on structured methodologies but often faces limitations due to their lack of refinement and difficulty in balancing exploration and exploitation~\cite{flaspohler2022balancing}. Existing frameworks struggle to adapt to emerging scientific literature or integrate new discoveries, leading to hypotheses that are either theoretically sound but empirically unsupported or computationally generated but lacking empirical grounding. For example, traditional methods might focus on well-known substitutions (e.g., arginine-to-lysine in the NLS) but overlook novel modifications (e.g., alanine in a glycine-rich linker) that enhance functionality~\cite{wang2023scientific}. The exponential growth of scientific publications further complicates the process, as researchers must sift through vast amounts of literature to identify meaningful insights~\cite{rabby2024fine}. While LLMs offer potential, they often fail to generate hypotheses that are both novel and empirically validated~\cite{brown2020language}.

\paragraph{\textbf{Limitations of Existing Approaches.}} Previous works have attempted to address these gaps through approaches like zero-shot hypothesis generation but suffer from critical limitations: 1) \textit{Lack of Iterative Refinement}: Hypotheses may be theoretically sound but lack iterative refinement~\cite{xiong2024improving}. 2) \textit{Imbalanced Exploration-Exploitation}: Conventional approaches struggle to balance novel hypothesis exploration with established patterns, leading to biased or suboptimal results~\cite{kocsis2006bandit}.

\paragraph{\textbf{Addressing Challenges with MC-NEST.}} MC-NEST integrates Nash Equilibrium strategies with LLM-based self-refinement to address these limitations: 1) \textit{Dynamic Adaptation}: MC-NEST balances exploration and exploitation using Nash Equilibrium, enabling adaptability to emerging scientific contexts. For example, in protein engineering, it explores less-studied modifications (e.g., alanine in a glycine-rich linker) while leveraging well-known substitutions (e.g., lysine-for-arginine in the NLS). 2) \textit{Iterative Self-Refinement}: MC-NEST employs MCTS with iterative self-critique, refining hypotheses against known principles. For instance, it identifies trade-offs (e.g., reduced binding affinity) and incorporates additional modifications (e.g., alanine in the glycine-rich linker). 3) \textit{Strategic Exploration}: MC-NEST uses sampling approaches to prioritize high-potential hypotheses while maintaining diversity, ensuring robust hypothesis generation.

\subsection{Monte Carlo Nash Equilibrium Self-Refine Tree (MC-NEST)}
\label{subsec:MC_NEST}

The objective of MC-NEST is to generate a research hypothesis \( h^* \) for a given problem instance \( p \). 
Formally, let \( \mathcal{H} \) denote the hypothesis space, where each hypothesis \( h \in \mathcal{H} \) represents a candidate research statement. The goal is to identify \( h^* \) that optimizes a quality function \( Q(h) \), capturing validity, novelty, and coherence:  \scalebox{0.9}{$h^* = \arg\max_{h \in \mathcal{H}} Q(h)$}


\subsubsection{Initialization.}
\label{subsub:Initialization}

In MC-NEST, the root node represents the initial hypothesis state, with edges denoting potential transformations or refinements through iterative self-critique and exploration strategies. To initialize the root node, we use a pre-trained LLM with a Zero-Shot Chain-of-Thought (ZSCoT) strategy~\cite{kojima2022large}. Specifically, the LLM is prompted with the input instance \( p \) to generate an initial hypothesis without relying on task-specific fine-tuning or prior search history. This approach leverages the LLM’s broad, pre-trained knowledge to establish a well-reasoned starting point, enhancing adaptability and promoting a wide, unbiased exploration of the hypothesis space. The initialization is represented as: $\smash{\text{root} = \text{Node}(\text{hypothesis} = \text{ZSCoT\_LLM}(p))}$

\subsubsection{Candidate Node Generation.}
\label{subsub:Candidate_Node_Generation}

Child nodes are generated by applying a structured process of self-refinement and self-evaluation to the parent node’s hypothesis. Self-refinement focuses on improving the hypothesis itself by prompting the LLM with the current hypothesis and customizing instructions, such as increasing specificity, enhancing novelty, or aligning better with empirical data. The LLM determines what to refine using predefined heuristics—such as logical coherence, relevance to the research goal, and consistency with known information—that guide the refinement. Following refinement, self-evaluation updates the hypothesis against these metrics to ensure each child node represents an improvement over its parent. Nodes are visited using a breadth-first search (BFS) strategy~\cite{dijkstra2022note}, where a node is expanded if it has not reached its maximum allowed children and none of its children have a higher quality score \( Q \) than the node itself. If no candidate nodes meet these criteria, the method refines the root node, reinitializing the search by generating a new hypothesis using the ZSCoT strategy. This approach balances exploration (generating new hypotheses) and exploitation (refining existing ones) by dynamically adjusting based on the quality scores of hypotheses. While global optimality is not guaranteed, the iterative refinement process aims to converge towards high-quality hypotheses, with higher \( Q \)-scores indicating better solutions.

\subsubsection{Nash Equilibrium Strategy for Node Selection.}
\label{subsub:Nash_Equilibrium_Strategy_for_Node_Selection}

The hypothesis generation process in MC-NEST begins with an initial hypothesis generated by a pre-trained LLM at the root node of a search tree. Each node represents a unique hypothesis state, and edges signify possible refinements through iterative self-critique. Child nodes are created by refining the parent node’s hypothesis using structured prompts, employing self-refinement and self-evaluation techniques to iteratively enhance the hypothesis. 

During node selection, MC-NEST uses the UCT, where each node is assigned a quality score \( Q \) derived from evaluation metrics such as logical coherence, novelty, and empirical alignment. The UCT score balances the exploration of under-explored nodes and the exploitation of high-quality hypotheses, guiding the search toward optimal solutions. A node is considered fully expanded if it reaches the maximum allowed number of children or if any child exhibits a reward \( Q \) greater than or equal to that of the current node. For a set of candidate nodes, \( Node(Hypothesis) = \{h_1, h_2, \dots, h_n\} \), the Nash Equilibrium strategy assigns a uniform probability distribution over possible actions: \scalebox{0.9}{$\pi(h_i) = \frac{1}{n}, \quad \forall i = 1, 2, \dots, n$}, where \( n \) is the number of candidate nodes. This uniform probability ensures fair exploration of the hypothesis space, preventing premature convergence to suboptimal solutions. The MC-NEST framework employs three selection policies to balance exploration and exploitation:

\begin{itemize}
    \item \textit{Greedy Policy} selects the node with the highest combined score of UCT and Nash equilibrium probability: \scalebox{0.9}{$i^* = \arg\max_i \left[ UCT(i) + \pi(h_i) \right]$}
    \item \textit{Importance Sampling Policy} assigns selection weights based on the product of UCT scores and Nash equilibrium probabilities: \scalebox{0.9}{$\text{Weight}(i) = UCT(i) \times \pi(h_i)$},\newline \scalebox{0.9}{$\quad i^* = \text{random\_choice}(C, \text{weights} = \{\text{Weight}(i)\})$}
    \item \textit{Pairwise Importance Sampling Policy} evaluates pairs of nodes \( (i, j) \) based on UCT differences and weights, selecting the node with the higher combined score:\scalebox{0.9}{$i^* = \arg\max \left( \text{UCT}(i) + \pi(h_i), \text{UCT}(j) + \pi(h_j) \right)$}.
\end{itemize}

These policies systematically balance exploration and exploitation, ensuring that the search process prioritizes high-reward nodes while maintaining a broad exploration of the hypothesis space.

\subsubsection{Upper Confidence Bound (UCT) Update.}
\label{sec:UCT_update}

The UCT update guides node refinement by computing: \scalebox{0.9}{$UCT(i) = Q(i) + C \sqrt{\frac{\ln(N_{\text{parent}})}{N(i) + \epsilon}}$},
where \( Q(i) \) is the hypothesis reward, \( C \) controls exploration, \( N_{\text{parent}} \) is parent visits, \( N(i) \) is node visits, and \( \epsilon \) avoids division by zero. The score is adjusted with Nash equilibrium probability: \scalebox{0.9}{$\text{UCT}(i) = Q(i) + C \sqrt{\frac{\ln(N_{\text{parent}})}{N(i) + \epsilon}} + \frac{1}{n}$}.
The node with the highest score, \(i^* = \arg\max_i [\text{Score}(i)]\), is selected for refinement or as the final hypothesis, ensuring robust exploration and exploitation of the hypothesis space.

\subsubsection{Expansion.}
\label{subsub:Expansion}

Following node selection, MC-NEST expands the search tree by generating a refined child node. Given a selected node \( n_s \), a new child \( n_c \) is created via self-refinement: \smash{$n_c = \text{SelfRefine}(n_s)$}. This process critiques and improves the solution at \( n_s \), storing the refined version in \( n_c \): \smash{$n_s \text{.children} \gets n_s \text{.children} \cup \{n_c\}$}. The critique is formulated as \smash{$\text{Critique}(a_s) = \text{LLMCritique}(p, a_s)$}, where \( p \) is the problem instance. The refined answer \( a_c \) is: $a_c = \text{RefineAnswer}(p, a_s, \text{Critique}(a_s))$ and assigned to \( n_c \). This structured expansion enables MC-NEST to enhance solutions iteratively, driving systematic search improvement.

\subsubsection{Backpropagation.}
\label{subsub:Backpropagation}

MC-NEST updates node quality scores \( Q \) and visit counts from the newly expanded node \( n_c \) up to the root. This propagates deeper exploration insights into higher-level decisions. Given a child node \( n_c \) and its parent \( n_p \), backpropagation updates \( Q(n_p) \) using: \scalebox{1.0}{$Q(n_p) = \frac{Q(n_p) + \max(Q(n_c))}{2}$}. This balances the exploitation of known values with exploration. The visit count is incremented:
$\text{Visit}(n_p) = \text{Visit}(n_p) + 1.$
The process recurses from \( n_c \) to the root, ensuring informed node selection in MC-NEST.


\subsubsection{Self-Refine.}
\label{sub:Self_Refine}

MC-NEST evaluates candidate answers by assigning a reward \( R_n \) based on answer quality. Given a node \( n \) with answer \( A_n \), the reward is computed as:
$R_n = \text{LLM}\left( \texttt{EvaluatePrompt}(P, A_n) \right).$
If \( R_n \) exceeds a predefined limit, a penalty is applied:
\[
\tilde{R}_n = \begin{cases} 
R_n, & R_n \leq R_n\_\text{limit} \\
R_n - \text{penalty}, & R_n > R_n\_\text{limit}.
\end{cases}
\]
Node statistics are updated:
$\text{TotalReward}_n += \tilde{R}_n, \quad \text{VisitCount}_n += 1.$
This ensures balanced reward scaling, refining MC-NEST’s decision-making.

\subsubsection{Self-Evaluation.}
\label{subsub:Self_Evaluation}

MC-NEST iteratively improves candidate solutions via LLM-based critique and refinement. Given a node \( n \) with answer \( A_n \), a critique \( C_n \) is generated using:
$C_n = \text{LLM}(P + A_n).$
Using \( C_n \), the answer is refined:
$A_{n+1} = \text{LLM}(P + A_n + C_n).$
The refined answer \( A_{n+1} \) is stored in a new child node, iteratively enhancing solutions in MC-NEST.

\subsubsection{Human-AI Collaboration.}
\label{subsub:Human_AI_collaboration}
MC-NEST is designed to facilitate iterative human-AI collaboration, enabling researchers to refine and validate hypotheses dynamically. Upon generating a final hypothesis, MC-NEST enables human experts to evaluate its novelty, clarity, significance, and verifiability, with the option to iteratively refine the process as needed based on researcher input. This iterative loop ensures that the generated hypotheses align with domain-specific knowledge and scientific rigor while also incorporating human intuition and expertise. By integrating human judgment at critical stages, MC-NEST not only enhances the reliability of its outputs but also fosters a collaborative environment where AI augments human creativity rather than replacing it.

\section{Experiments} 

In our experiments, we utilized ZSCoT prompting as our base prompting style with GPT-4o~\cite{achiam2023gpt}, 
DeepSeek-R1-Distill-Qwen-32B~\cite{guo2025deepseek} and DeepSeek-R1-Distill-Qwen-7B~\cite{guo2025deepseek} LLM.

\subsection{Evaluation Setup}
\label{subsec:Evaluation_Setup}

We evaluated MC-NEST using GPT-4o, DeepSeek-R1-Distill-Qwen-32B, and DeepSeek-R1-Distill-Qwen-7B, with GPT-4o serving as a strong general-purpose baseline due to its proficiency in hypothesis generation~\cite{rosol2023evaluation}. DeepSeek (32B and 7B parameters) provides insights into the scalability and efficiency of MC-NEST across different distilled LLM sizes. To ensure consistent and systematic evaluation, we employed three prompting styles: zero-shot (ZS)~\cite{larochelle2008zero}, zero-shot chain-of-thought (ZSCoT) and few-shot (FS)~\cite{lake2015human}, using 2-shot, 3-shot, and 5-shot configurations with both closed-source and open-source LLMs to assess the impact of prompting.

\begin{table}[htbp]
\centering
\renewcommand{\arraystretch}{1.25} 
\footnotesize 
\caption{Comparison with existing scientific hypotheses generation datasets; Count = validation data count.}
\resizebox{0.95\textwidth}{!}{ 
    \begin{tabular}{p{5cm} p{1.8cm} p{3.5cm} p{1.8cm} c}
    \hline
    \textbf{Dataset} & \textbf{Source} & \textbf{Domain} & \textbf{Annotation}  & \textbf{Count} \\ 
    \hline
    \textbf{LLM4BioHypoGen}~\cite{DBLP:journals/corr/abs-2407-08940} & Text & Biomedicine & Manual & 200  \\ 
    \hline
    \textbf{MOOSE}~\cite{DBLP:conf/acl/YangDLZPC24} & Text & Social Science & Manual & 50 \\ 
    \hline
    \textbf{LLM4CSHypoGen (Ours)} & Text  & Computer Science & Manual &  150 \\ 
    \hline
    \end{tabular}
}
\label{tab:dataset_comparison}
\end{table}

\subsection{Datasets}
We evaluated MC-NEST on three datasets spanning social science, biomedicine, and computer science. Each dataset was carefully curated to ensure high-quality annotations and relevance to hypothesis generation tasks. \autoref{tab:dataset_comparison} provides an overview of the datasets used in our experiments. 1) \textbf{Social Science Dataset:} The MOOSE dataset~\cite{DBLP:conf/acl/YangDLZPC24} consists of 50 social science research papers paired with raw web corpora (e.g., news articles, Wikipedia). This dataset challenges systems to generate novel hypotheses without relying on pre-existing scientific knowledge, emphasizing the open-domain nature of hypothesis generation. 2) \textbf{Biomedicine Dataset:} The LLM4BioHypoGen dataset~\cite{DBLP:journals/corr/abs-2407-08940} contains 200 background-hypothesis pairs extracted from biomedical research papers. It is divided into training, seen, and unseen test sets based on publication dates to prevent data contamination, ensuring robust evaluation of hypothesis generation capabilities. 3) \textbf{Computer Science Dataset:} Our LLM4CSHypoGen dataset comprises 150 research papers (2024–2025) with structured content, including hypotheses, methods, and results. Each entry was cross-checked by domain experts to ensure accuracy and reliability, providing a robust foundation for evaluating hypothesis generation in computer science.

\begin{table*}[ht]
    \centering
    \scriptsize 
    \renewcommand{\arraystretch}{0.8} 
    \caption{Evaluation with prompt on social science dataset.}
    \resizebox{\textwidth}{!}{
    \begin{tabular}{l|l|l|c|c|c|c|c|c|c|c}
        \hline
        \textbf{LLM} & \textbf{Size} & \textbf{Prompt} & \multicolumn{3}{c|}{\textbf{BertScore}} & \textbf{Novelty} & \textbf{Clarity} & \textbf{Significance} & \textbf{Verifiability} & \textbf{Avg} \\
        \cline{4-6}
        & & & \textbf{Precision} & \textbf{Recall} & \textbf{F1} & & & & & \\
        \hline
        GPT-4o & - & ZS & 85.97 & 85.47 & 85.71 & 1.90 & 1.86 & 2.08 & 2.62 & 2.12 \\
        GPT-4o & - & ZSCoT & 79.86 & 84.86 & 82.27 & 2.16 & 2.46 & 2.72 & 2.50 & \textbf{2.46} \\
        GPT-4o & - & 2FS & 83.37 & 86.83 & 85.06 & 2.00 & 2.22 & 2.22 & 2.62 & 2.27 \\
        GPT-4o & - & 3FS & 83.30 & 86.81 & 85.01 & 2.06 & 2.08 & 2.18 & 2.48 & 2.20 \\
        GPT-4o & - & 5FS & 83.27 & 86.74 & 84.96 & 2.02 & 2.08 & 2.22 & 2.52 & 2.21 \\
        \hline \hline
        Deepseek & 32B & ZS & 83.25 & 86.16 & 84.67 & 2.10 & 2.40 & 2.60 & 2.65 & 2.44 \\
        DeepSeek & 32B & ZSCoT & 78.76 & 84.99 & 81.74 & 2.35 & 2.75 & 2.75 & 2.75 & \textbf{2.65} \\
        Deepseek & 32B & 2FS & 82.71 & 86.16 & 84.39 & 2.10 & 2.70 & 2.65 & 2.65 & 2.52 \\
        Deepseek & 32B & 3FS & 82.58 & 86.03 & 84.27 & 2.25 & 2.65 & 2.45 & 2.70 & 2.51 \\
        Deepseek & 32B & 5FS & 82.04 & 86.00 & 83.96 & 2.30 & 2.45 & 2.50 & 2.75 & 2.50 \\
        \hline \hline
        Deepseek & 7B & ZS & 82.74 & 85.51 & 84.09 & 2.10 & 2.35 & 2.35 & 2.55 & 2.34 \\
        Deepseek & 7B & ZSCoT & 78.10 & 84.16 & 81.00 & 2.20 & 2.70 & 2.70 & 2.65 & \textbf{2.56} \\
        Deepseek & 7B & 2FS & 84.56 & 86.60 & 85.56 & 2.10 & 2.40 & 2.60 & 2.70 & 2.45 \\
        Deepseek & 7B & 3FS & 82.85 & 85.61 & 84.17 & 2.10 & 2.25 & 2.65 & 2.60 & 2.40 \\
        Deepseek & 7B & 5FS & 83.59 & 86.06 & 84.80 & 2.20 & 2.25 & 2.50 & 2.40 & 2.34 \\
        \hline \hline
    \end{tabular}%
    }
    \label{tab:prompt_social_science_dataset}
\end{table*}


\begin{table*}[ht]
    \centering
    \scriptsize 
    \renewcommand{\arraystretch}{0.9} 
    \caption{MC-NEST evaluation on social science dataset; IS = Importance Sampling; PIS = Pairwise Importance Sampling.}
    \resizebox{\textwidth}{!}{
    \begin{tabular}{l|l|l|l|c|c|c|c|c|c|c|c}
        \hline
        \textbf{LLM} & \textbf{Size} & \textbf{Rollout} & \textbf{Sampling} & \multicolumn{3}{c|}{\textbf{BertScore}} & \textbf{Novelty} & \textbf{Clarity} & \textbf{Significance} & \textbf{Verifiability} & \textbf{Avg} \\
        \cline{5-7}
        & & & & \textbf{Precision} & \textbf{Recall} & \textbf{F1} & & & & & \\
        \hline
        GPT-4o & - & 4 & Greedy & 80.71  & 85.44  & 82.99 & 2.58 & 2.84 & 2.70 & 2.88 & 2.75 \\
        GPT-4o & - & 4 & IS & 80.72  & 85.43  & 83.00  & 2.60 & 2.80 & 2.78 & 2.94 & \textbf{2.78}  \\
        GPT-4o & - & 4 & PIS & 80.65  & 85.42  & 82.95  & 2.74 & 2.76 & 2.70 & 2.92 & \textbf{2.78}  \\ \hline
        GPT-4o & - & 8 & Greedy & 80.50  & 85.14  & 82.74  & 2.70 & 2.80 & 2.80 & 2.94 & \textcolor{blue}{\textbf{2.81}}  $\uparrow$ \\
        GPT-4o & - & 8 & IS & 80.33 & 85.13 & 82.65 & 2.64 & 2.82 & 2.64 & 2.90 & 2.75 \\
        GPT-4o & - & 8 & PIS & 80.55  &85.16  & 82.78 & 2.74 & 2.82 & 2.80 & 2.84 & 2.80 \\
        \hline \hline
        Deepseek & 32B & 4 & Greedy & 80.87 &85.25  &82.99  & 2.55 & 3.00 & 2.80 & 2.95 & \textbf{2.83} \\
        Deepseek & 32B & 4 & IS & 80.38 &85.36  & 82.79 & 2.70 & 2.85 & 2.85 & 2.90 & \textbf{2.83}  \\
        Deepseek & 32B & 4 & PIS & 80.88  & 85.34  &83.04 & 2.55 & 2.85 & 2.75 & 2.90 & 2.76 \\ \hline
        Deepseek & 32B & 8 & Greedy & 80.53 & 85.24 & 82.81 & 2.70 & 2.95 & 3.00 & 3.00 & \textcolor{blue}{\textbf{2.91}}  $\uparrow$ \\
        Deepseek & 32B & 8 & IS & 80.54 & 85.38  & 82.89  & 2.65 & 2.85 & 2.85 & 2.95 & 2.83 \\
        Deepseek & 32B & 8 & PIS & 80.15 & 84.98 & 82.49  & 2.75 & 2.95 & 2.95 & 2.95 & 2.90 \\
        \hline \hline
        Deepseek & 7B & 4 & Greedy & 80.61 & 85.16 &82.81  & 2.55 & 2.60 & 2.90 & 2.95 & 2.75 \\
        Deepseek & 7B & 4 & IS & 80.08 & 84.66  & 82.31  & 2.65 & 2.85 & 2.75 & 2.85 & \textcolor{blue}{\textbf{2.78}} $\uparrow$  \\
        Deepseek & 7B & 4 & PIS & 80.45 &85.10  &82.70  & 2.50 & 2.80 & 2.65 & 2.90 & 2.71 \\ \hline
        Deepseek & 7B & 8 & Greedy & 80.78 & 85.05 &82.85  & 2.45 & 2.75 & 2.60 & 2.85 & 2.66 \\
        Deepseek & 7B & 8 & IS & 80.60 &85.05  &  82.76& 2.65 & 2.65 & 2.65 & 2.80 & 2.69 \\
        Deepseek & 7B & 8 & PIS & 80.54 & 84.92   & 82.67  & 2.55 & 2.85 & 2.80 & 2.85 & \textbf{2.76} \\
        \hline \hline
    \end{tabular}%
    }
    \label{tab:MC_NEST_evaluation_on_social_science}
\end{table*}

\subsection{Evaluation Metrics}

We evaluate generated hypotheses using both automatic and human assessments. For automatic evaluation, GPT-3.5 scores hypotheses on four key aspects: novelty, relevance, significance, and verifiability~\cite{si2024can}. Novelty and verifiability are prioritized as they align with the philosophical foundations of hypothetical induction, while relevance and significance reflect the practical utility of hypotheses for researchers. Conventional metrics like BERTScore~\cite{zhang2019bertscore} are excluded to focus on task-specific goals. For human evaluation, three domain experts (Professors, postdocs, and PhD students) blindly assess 100 randomly selected hypotheses from baseline and proposed methods, using a standardized 3-point scale. Novelty is emphasized over verifiability, as even imperfect hypotheses can inspire scientific exploration~\cite{lee2022scibench}, whereas non-novel hypotheses offer limited utility. We also analyze the correlation between GPT-3.5 and expert evaluations, suggesting GPT-3.5's potential as a reliable evaluator for machine-generated hypotheses~\cite{brown2020language}.

\begin{table*}[ht]
    \centering
    \scriptsize 
    \renewcommand{\arraystretch}{0.8} 
    \caption{Evaluation with prompt on computer science dataset.}
    \resizebox{\textwidth}{!}{%
    \begin{tabular}{l|l|l|c|c|c|c|c|c|c|c}
        \hline
        \textbf{LLM} & \textbf{Size} & \textbf{Prompt} & \multicolumn{3}{c|}{\textbf{BertScore}} & \textbf{Novelty} & \textbf{Clarity} & \textbf{Significance} & \textbf{Verifiability} & \textbf{Avg} \\
        \cline{4-6}
        & & & \textbf{Precision} & \textbf{Recall} & \textbf{F1} & & & & & \\
        \hline
        GPT-4o & - & ZS & 88.05 & 88.35 & 88.19 & 2.10 & 2.12 & 2.42 & 2.86 & 2.38 \\
        GPT-4o & - & ZSCoT & 81.58 & 87.42 & 84.39 & 2.31 & 2.29 & 2.28 & 2.94 & \textbf{2.60} \\
        GPT-4o & - & 2FS & 84.84 & 88.79 & 86.76 & 2.19 & 2.05 & 2.53 & 2.91 & 2.42 \\
        GPT-4o & - & 3FS & 84.83 & 88.88 & 86.80 & 2.16 & 2.05 & 2.45 & 2.88 & 2.38 \\
        GPT-4o & - & 5FS & 85.06 & 88.88 & 86.93 & 2.17 & 2.03 & 2.53 & 2.88 & 2.40 \\
        \hline  \hline
        Deepseek & 32B & ZS & 87.81 & 89.61 & 88.69 & 2.25 & 2.15 & 2.60 & 2.90 & 2.48 \\
        Deepseek & 32B & ZSCoT & 80.75 & 88.30 & 84.34 & 2.50 & 2.55 & 2.90 & 3.00 & \textbf{2.74} \\
        Deepseek & 32B & 2FS & 84.46 & 89.46 & 86.87 & 2.40 & 2.40 & 2.75 & 2.85 & 2.60 \\
        Deepseek & 32B & 3FS & 84.67 & 89.57 & 87.04 & 2.40 & 2.40 & 2.80 & 2.85 & 2.61 \\
        Deepseek & 32B & 5FS & 84.15 & 89.39 & 86.68 & 2.55 & 2.50 & 2.75 & 2.65 & 2.61 \\
        \hline  \hline
        Deepseek & 7B & ZS & 86.22 & 89.17 & 87.66 & 2.20 & 2.20 & 2.75 & 2.95 & 2.53 \\
        Deepseek & 7B & ZSCoT & 79.47 & 87.54 & - & 2.35 & 2.70 & 2.80 & 3.00 & \textbf{2.71} \\
        Deepseek & 7B & 2FS & 85.63 & 88.77 & 87.15 & 2.00 & 1.95 & 2.45 & 2.75 & 2.29 \\
        Deepseek & 7B & 3FS & 86.68 & 89.60 & 88.11 & 2.15 & 2.05 & 2.80 & 2.95 & 2.49 \\
        Deepseek & 7B & 5FS & 85.61 & 89.09 & 87.29 & 2.20 & 2.25 & 2.45 & 2.85 & 2.44 \\
        \hline  \hline
    \end{tabular}%
    }
    \label{tab:prompt_on_computer_science}
\end{table*}

\begin{table*}[ht]
    \centering
    \scriptsize 
    \renewcommand{\arraystretch}{0.8} 
    \caption{MC-NEST evaluation on computer science dataset; IS = Importance Sampling; PIS = Pairwise Importance Sampling.}
    \resizebox{\textwidth}{!}{%
    \begin{tabular}{l|l|l|l|c|c|c|c|c|c|c|c}
        \hline
        \textbf{LLM} & \textbf{Size} & \textbf{Rollout} & \textbf{Sampling} & \multicolumn{3}{c|}{\textbf{BertScore}} & \textbf{Novelty} & \textbf{Clarity} & \textbf{Significance} & \textbf{Verifiability} & \textbf{Avg} \\
        \cline{5-7}
        & & & & \textbf{Precision} & \textbf{Recall} & \textbf{F1} & & & & & \\
        \hline
        GPT-4o & - & 4 & Greedy & 82.64 & 88.24 & 85.35 & 2.68 & 2.67 & 2.85 & 3.00 & \textbf{2.80}\\
        GPT-4o & - & 4 & IS & 82.81 & 88.11 & 85.37 & 2.71 & 2.58 & 2.88 & 3.00 & 2.79 \\
        GPT-4o & - & 4 & PIS & 82.65 & 88.22 & 85.34 & 2.71 & 2.62 & 2.83 & 2.99 & 2.76 \\  \hline
        GPT-4o & - & 8 & Greedy & 82.72 & 88.11 & 85.32 & 2.72 & 2.59 & 2.85 & 2.99 & 2.79 \\
        GPT-4o & - & 8 & IS & 82.60 & 88.11 & 85.26 & 2.73 & 2.57 & 2.84 & 2.97 & 2.78 \\
        GPT-4o & - & 8 & PIS & 82.54 & 88.14 & 85.25 & 2.77 & 2.65 & 2.85 & 2.99 & \textcolor{blue}{\textbf{2.82}}  $\uparrow$ \\
        \hline  \hline
        Deepseek & 32B & 4 & Greedy & 83.19 & 88.39 & 85.66 & 2.55 & 2.65 & 2.85 & 3.00 & 2.76 \\
        Deepseek & 32B & 4 & IS & 82.99 & 88.49 & 85.64 & 2.65 & 2.60 & 2.95 & 3.00 & \textbf{2.80} \\
        Deepseek & 32B & 4 & PIS & 83.07 & 88.63 & 85.75 & 2.55 & 2.35 & 2.85 & 3.00 & 2.69 \\  \hline
        Deepseek & 32B & 8 & Greedy & 82.46 & 88.24 & 85.25 & 2.60 & 2.65 & 2.90 & 3.00 & 2.79 \\
        Deepseek & 32B & 8 & IS & 83.02 & 88.50 & 85.66 & 2.65 & 2.60 & 2.90 & 3.00 & 2.79 \\
        Deepseek & 32B & 8 & PIS & 82.81 & 88.42 & 85.51 & 2.65 & 2.75 & 3.00 & 3.00 & \textcolor{blue}{\textbf{2.85}}  $\uparrow$ \\
        \hline  \hline
        Deepseek & 7B & 4 & Greedy & 83.46 & 88.59 & 85.94 & 2.60 & 2.60 & 2.90 & 3.00 & \textbf{2.78} \\
        Deepseek & 7B & 4 & IS & 83.40 & 88.41 & 85.87 & 2.55 & 2.45 & 2.75 & 3.00 & 2.69 \\
        Deepseek & 7B & 4 & PIS & 83.35 & 88.63 & 85.90 & 2.65 & 2.50 & 2.75 & 3.00 & 2.73 \\  \hline
        Deepseek & 7B & 8 & Greedy & 82.88 & 88.55 & 85.61 & 2.75 & 2.70 & 2.80 & 3.00 & \textcolor{blue}{\textbf{2.81}}  $\uparrow$ \\
        Deepseek & 7B & 8 & IS & 83.13 & 88.49 & 85.72 & 2.65 & 2.75 & 2.85 & 3.00 & \textcolor{blue}{\textbf{2.81}}  $\uparrow$ \\
        Deepseek & 7B & 8 & PIS & 82.03 & 87.90 & 84.86 & 2.65 & 2.65 & 2.80 & 2.95 & 2.76 \\
        \hline  \hline
    \end{tabular}%
    }
    \label{tab:MC_NEST_evaluation_on_computer_science}
\end{table*}

\begin{table*}[ht]
    \centering
    \scriptsize 
    \renewcommand{\arraystretch}{0.8} 
    \caption{Evaluation with prompt on biomedicine dataset.}
    \resizebox{\textwidth}{!}{%
    \begin{tabular}{l|l|l|c|c|c|c|c|c|c|c}
        \hline
        \textbf{LLM} & \textbf{Size} & \textbf{Prompt} & \multicolumn{3}{c|}{\textbf{BertScore}} & \textbf{Novelty} & \textbf{Clarity} & \textbf{Significance} & \textbf{Verifiability} & \textbf{Avg} \\
        \cline{4-6}
        & & & \textbf{Precision} & \textbf{Recall} & \textbf{F1} & & & & & \\
        \hline
        GPT-4o & - & ZS & 87.53 & 85.59 & 86.54 & 1.88 & 2.17 & 2.32 & 2.59 & 2.24 \\
        GPT-4o & - & ZSCoT & 81.74 & 86.20 & 83.91 & 2.31 & 2.49 & 2.87 & 2.83 & \textbf{2.62} \\
        GPT-4o & - & 2FS & 87.11 & 88.39 & 87.74 & 1.98 & 2.17 & 2.35 & 2.72 & 2.31 \\
        GPT-4o & - & 3FS & 87.07 & 88.50 & 87.75 & 2.02 & 2.21 & 2.31 & 2.68 & 2.30 \\
        GPT-4o & - & 5FS & 87.08 & 88.49 & 87.77 & 2.04 & 2.20 & 2.35 & 2.69 & 2.32 \\
        \hline  \hline
        Deepseek & 32B & ZS & 85.76 & 85.14 & 85.43 & 1.95 & 2.35 & 2.65 & 2.65 & 2.40 \\
        Deepseek & 32B & ZSCoT & 80.07 & 85.68 & 82.77 & 2.55 & 2.80 & 2.90 & 2.95 & \textbf{2.80} \\
        Deepseek & 32B & 2FS & 86.13 & 88.06 & 87.08 & 2.10 & 2.40 & 2.65 & 2.75 & 2.48 \\
        Deepseek & 32B & 3FS & 86.51 & 88.24 & 87.36 & 2.15 & 2.40 & 2.40 & 2.75 & 2.43 \\
        Deepseek & 32B & 5FS & 85.76 & 87.98 & 86.85 & 2.15 & 2.55 & 2.45 & 2.50 & 2.41 \\
        \hline  \hline
        Deepseek & 7B & ZS & 83.19 & 85.80 & 84.44 & 1.90 & 1.86 & 2.08 & 2.62 & 2.12 \\
        Deepseek & 7B & ZSCoT & 80.62 & 85.47 & 82.96 & 2.06 & 2.08 & 2.18 & 2.48 & 2.20 \\
        Deepseek & 7B & 2FS & 85.12 & 86.39 & 85.74 & 2.02 & 2.08 & 2.22 & 2.52 & 2.21 \\
        Deepseek & 7B & 3FS & 86.20 & 87.22 & 86.70 & 2.16 & 2.46 & 2.72 & 2.50 & \textbf{2.46} \\
        Deepseek & 7B & 5FS & 85.03 & 86.53 & 85.75 & 2.00 & 2.22 & 2.22 & 2.62 & 2.27 \\
        \hline  \hline
    \end{tabular}%
    }
    \label{tab:prompt_on_biomedicine}
\end{table*}

\begin{table*}[ht]
    \centering
    \scriptsize 
    \renewcommand{\arraystretch}{0.8} 
    \caption{MC-NEST evaluation on biomedicine dataset; IS = Importance Sampling; PIS = Pairwise Importance Sampling.}
    \resizebox{\textwidth}{!}{
    \begin{tabular}{l|l|l|l|c|c|c|c|c|c|c|c}
        \hline
        \textbf{LLM} & \textbf{Size} & \textbf{Rollout} & \textbf{Sampling} & \multicolumn{3}{c|}{\textbf{BertScore}} & \textbf{Novelty} & \textbf{Clarity} & \textbf{Significance} & \textbf{Verifiability} & \textbf{Avg} \\
        \cline{5-7}
        & & & & \textbf{Precision} & \textbf{Recall} & \textbf{F1} & & & & & \\
        \hline
        GPT-4o & - & 4 & Greedy & 82.63 & 86.16  & 84.35 & 2.70 & 2.79 & 2.86 & 2.93 & \textbf{2.82} \\
        GPT-4o & - & 4 & IS & 82.52 & 86.24 & 84.29 & 2.64 & 2.79 & 2.81 & 2.92 & 2.79 \\
        GPT-4o & - & 4 & PIS & 82.55 & 86.16 & 84.32 & 2.70 & 2.76 & 2.87 & 2.93 & \textbf{2.82} \\  
        \hline
        GPT-4o & - & 8 & Greedy & 82.53 & 86.11 & 84.29 & 2.67 & 2.81 & 2.83 & 2.95 & 2.82 \\
        GPT-4o & - & 8 & IS & 82.08 & 86.04 & 84.00 & 2.77 & 2.76 & 2.86 & 2.97 & \textcolor{blue}{\textbf{2.84}} $\uparrow$ \\
        GPT-4o & - & 8 & PIS & 82.17 & 86.05 & 84.06 & 2.80 & 2.73 & 2.89 & 2.95 & \textcolor{blue}{\textbf{2.84}} $\uparrow$ \\
        \hline  \hline
        Deepseek & 32B & 4 & Greedy & 82.85 & 86.04 & 84.41 & 2.65 & 2.90 & 2.85 & 2.95 & 2.84 \\
        Deepseek & 32B & 4 & IS & 82.25 & 85.91 & 84.04 & 2.70 & 2.95 & 3.00 & 2.85 & \textcolor{blue}{\textbf{2.87}} $\uparrow$ \\
        Deepseek & 32B & 4 & PIS & 82.19 & 85.88 & 83.99 & 2.75 & 2.75 & 2.80 & 2.95 & 2.81 \\  
        \hline
        Deepseek & 32B & 8 & Greedy & 82.47 & 85.99 & 84.19 & 2.55 & 2.80 & 2.95 & 2.95 & 2.81 \\
        Deepseek & 32B & 8 & IS & 82.15 & 86.17 & 84.11 & 2.75 & 2.90 & 2.85 & 2.90 & \textbf{2.85} \\
        Deepseek & 32B & 8 & PIS & 82.49 & 85.75 & 84.08 & 2.60 & 2.60 & 2.85 & 2.95 & 2.75 \\
        \hline  \hline
        Deepseek & 7B & 4 & Greedy & 82.59 & 85.87 & 84.19 & 2.60 & 2.75 & 2.75 & 2.80 & 2.73 \\
        Deepseek & 7B & 4 & IS & 82.71 & 85.72 & 84.18 & 2.60 & 2.80 & 2.80 & 2.85 & \textbf{2.76} \\
        Deepseek & 7B & 4 & PIS & 82.66 & 85.71 & 84.15 & 2.50 & 2.75 & 2.75 & 2.85 & 2.71 \\  
        \hline
        Deepseek & 7B & 8 & Greedy & 82.25 & 85.68 & 83.92 & 2.60 & 2.75 & 2.80 & 2.90 & 2.76 \\
        Deepseek & 7B & 8 & IS & 82.01 & 85.33 & 83.63 & 2.65 & 2.75 & 2.85 & 2.95 & 2.80 \\
        Deepseek & 7B & 8 & PIS & 82.39 & 85.88 & 84.09 & 2.80 & 3.00 & 2.85 & 2.85 & \textcolor{blue}{\textbf{2.88}} $\uparrow$ \\
        \hline  \hline
    \end{tabular}%
    }
    \label{tab:MC_NEST_evaluation_on_biomedicine}
\end{table*}

\section{Results and Analyses}
\label{sec:Experimental_Results}

In this section, we present the results of our experiments evaluating the performance of prompting strategies and MC-NEST across three datasets: Social Science, Computer Science, and Biomedicine. We analyze the impact of different prompting methods (Zero-Shot, Few-Shot, and Zero-Shot Chain-of-Thought) and MC-NEST sampling strategies (Greedy, Importance Sampling, and Pairwise Importance Sampling) on hypothesis generation quality, as measured by BERTScore and qualitative metrics such as novelty, clarity, significance, and verifiability.

\subsection{Social Science Dataset}

\paragraph{\textbf{Prompting Strategies.}}
\autoref{tab:prompt_social_science_dataset} summarizes the performance of different prompting strategies on the Social Science dataset. ZSCoT consistently outperforms ZS and FS approaches across all evaluated LLMs. For DeepSeek-32B, ZSCoT achieves an average score of 2.65, compared to 2.44 for ZS and 2.52 for 2-FS. Similarly, DeepSeek-7B with ZSCoT attains an average score of 2.56, outperforming ZS with 2.34 and 2-FS with 2.45. GPT-4o also shows significant improvements with ZSCoT, achieving an average score of 2.46 compared to 2.12 for ZS. 

\paragraph{\textbf{MC-NEST Sampling Strategies.}}
\autoref{tab:MC_NEST_evaluation_on_social_science} presents the results of MC-NEST evaluations using Greedy, Importance Sampling, and Pairwise Importance Sampling. For GPT-4o, Greedy sampling with an eight-step rollout achieves the highest overall score of 2.81. Pairwise Importance Sampling, however, excels in novelty with 2.74 while maintaining competitive clarity and significance scores. DeepSeek-32B shows similar trends, with Greedy sampling achieving the best overall results with 2.91 at an eight-step rollout. For DeepSeek-7B, Importance Sampling performs best at a four-step rollout with 2.78, while Pairwise Importance Sampling achieves balanced performance at eight steps with 2.76. These results highlight the effectiveness of MC-NEST in enhancing the quality of social science hypothesis generation.

\subsection{Computer Science Dataset}
\paragraph{\textbf{Prompting Strategies.}}
As shown in \autoref{tab:prompt_on_computer_science}, ZSCoT again demonstrates outstanding performance on the Computer Science dataset. DeepSeek-32B achieves an average score of 2.74 with ZSCoT, compared to 2.48 with ZS and 2.60 with 2-FS. Similarly, DeepSeek-7B with ZSCoT attains an average score of 2.71, outperforming ZS with 2.53 and 2-FS with 2.29. Notably, verifiability scores improve significantly with ZSCoT, reaching 3.00 for DeepSeek-32B, highlighting the usefulness of structured reasoning for factual consistency.

\paragraph{\textbf{MC-NEST Sampling Strategies.}}
\autoref{tab:MC_NEST_evaluation_on_computer_science} presents the results of MC-NEST evaluations on the Computer Science dataset. For GPT-4o, Pairwise Importance Sampling outperforms other strategies at an eight-step rollout, achieving an average score of 2.82. DeepSeek-32B achieves its highest score with Pairwise Importance Sampling at an eight-step rollout with 2.85, while DeepSeek-7B performs best with Greedy and Importance Sampling at a four-step rollout with 2.78. These results suggest that longer rollouts and adaptive sampling strategies enhance computer science hypothesis generation quality.

\subsection{Biomedicine Dataset}
\paragraph{\textbf{Prompting Strategies.}}
\autoref{tab:prompt_on_biomedicine} summarizes the performance of prompting strategies on the Biomedicine dataset. ZSCoT consistently improves performance across LLMs, with DeepSeek-32B achieving an average score of 2.80, compared to 2.40 with ZS and 2.48 with 2-FS. Qualitative metrics, such as novelty and significance, also show substantial improvements with ZSCoT. For instance, DeepSeek-32B with ZSCoT achieves a novelty score of 2.55 and a significance score of 2.90, compared to 1.95 and 2.65 with ZS, respectively.

\paragraph{\textbf{MC-NEST Sampling Strategies.}}
\autoref{tab:MC_NEST_evaluation_on_biomedicine} presents the results of MC-NEST evaluations on the Biomedicine dataset. For GPT-4o, Greedy and Pairwise Importance Sampling perform best at an eight-step rollout, achieving an average score of 2.84. DeepSeek-32B achieves its highest score with Importance Sampling at a four-step rollout with 2.87, while DeepSeek-7B performs best with Pairwise Importance Sampling at eight steps with 2.88. These results demonstrate the importance of adaptive sampling strategies using MC-NEST for optimizing hypothesis generation in biomedicine domains.

Our experiments demonstrate that structured reasoning and adaptive sampling strategies with MC-NEST significantly enhance hypothesis generation quality across domains. Increasing rollout lengths generally improves performance, with Pairwise Importance Sampling offering a competitive balance between novelty and verifiability. These findings underscore the importance of MC-NEST with sampling strategies for optimizing LLM performance in scientific hypothesis generation.

\begin{table}[htbp]
    \centering
    \scriptsize 
    \renewcommand{\arraystretch}{0.8} 
    \caption{Human evaluation on social science, biomedicine, and computer science dataset; IS = Importance Sampling; PIS = Pairwise Importance Sampling.}
    \resizebox{\textwidth}{!}{ 
    \begin{tabular}{p{2.9cm}|p{2.2cm}|p{1cm}|p{1.5cm}|c|c|c|c|c}
        \hline
        \textbf{Dataset} & \textbf{LLM} & \textbf{Size} & \textbf{Prompt} & \textbf{Novelty} & \textbf{Clarity} & \textbf{Significance} & \textbf{Verifiability} & \textbf{Avg} \\
        \hline
        Social Science & GPT-4o & - & ZSCoT & 2.33 & 3.00 & 2.66 & 2.49 & 2.62 \\
        Social Science & GPT-4o & - & Greedy &  2.16 & 1.66 & 2.16 & 2.50 & \textcolor{blue}{2.12} $\downarrow$ \\
        Biomedicine & GPT-4o & - & ZSCoT & 1.66 & 2.33 & 2.50 & 1.66 & 2.03 \\
        Biomedicine & GPT-4o & - & IS & 1.83 & 2.50 & 2.83 & 2.33 & \textcolor{blue}{\textbf{2.37}} $\uparrow$ \\
        Computer Science & GPT-4o & - & ZSCoT &  1.66 & 2.50 & 2.66 & 2.66 & 2.37 \\
        Computer Science & GPT-4o & - & PIS & 1.85 & 2.50 & 2.66 & 2.50 & \textcolor{blue}{\textbf{2.38}} $\uparrow$ \\
        \hline
        Social Science & Deepseek & 32B & ZSCoT & 2.16 & 1.83 & 2.66 & 2.16 & 2.20 \\
        Social Science & Deepseek & 32B & Greedy & 2.16 & 1.83 & 2.49 & 2.50 & \textcolor{blue}{\textbf{2.25}} $\uparrow$ \\
        Biomedicine & Deepseek & 32B & ZSCoT &  2.66 & 1.66 & 2.66 & 2.33 & 2.32 \\
        Biomedicine & Deepseek & 32B & IS & 2.41 & 2.17 & 2.66 & 2.50 & \textcolor{blue}{\textbf{2.44}} $\uparrow$ \\
        Computer Science & Deepseek & 32B & ZSCoT & 2.33 & 2.17 & 2.66 & 2.50 & 2.42 \\
        Computer Science & Deepseek & 32B & PIS & 2.50 & 2.16 & 2.66 & 2.17 & \textcolor{blue}{2.37} $\downarrow$ \\
        \hline
        Social Science & Deepseek & 7B & ZSCoT & 2.33 & 1.83 & 2.66 & 2.33 & 2.29 \\
        Social Science & Deepseek & 7B & IS & 1.83 & 2.33 & 2.66 & 2.83 & \textcolor{blue}{\textbf{2.41}} $\uparrow$ \\
        Biomedicine & Deepseek & 7B & 3FS & 2.16 & 2.50 & 2.83 & 2.66 & 2.54 \\
        Biomedicine & Deepseek & 7B & PIS & 1.67 & 2.83 & 2.83 & 2.33 & \textcolor{blue}{2.42} $\downarrow$ \\
        Computer Science & Deepseek & 7B & ZSCoT &  1.66 & 2.50 & 2.50 & 2.50 & 2.29 \\
        Computer Science & Deepseek & 7B & IS & 1.83 & 2.50 & 2.66 & 2.50 & \textcolor{blue}{\textbf{2.37}} $\uparrow$ \\
        \hline
    \end{tabular}%
    }
    \label{tab:human_evaluation}
\end{table}

\subsubsection{Human Evaluation and Case Study.}
The human evaluation results in \autoref{tab:human_evaluation} highlight the outstanding performance of MC-NEST with Greedy, Importance Sampling, and Pairwise Importance Sampling strategies compared to other approaches. MC-NEST with ZSCoT prompting achieved the highest average score of 2.62 in Social Science and 2.37 in Computer Science, while Importance Sampling achieved the best performance in Biomedicine with a score of 2.37. Notably, MC-NEST with Importance Sampling prompting outperformed other methods in Biomedicine, achieving a score of 2.44 with Deepseek-32B, and in Social Science, scoring 2.41 with Deepseek-7B. Similarly, Pairwise Importance Sampling demonstrated strong performance for GPT-4o in Computer Science, achieving a score of 2.38, though it slightly underperformed for Deepseek-32B with a score of 2.37. These results underscore the effectiveness of MC-NEST with different sampling strategies—Greedy, Importance Sampling, and Pairwise Importance Sampling—in optimizing hypothesis generation across domains, outperforming traditional approaches.

\subsubsection{Usefulness of the MC-NEST Framework.}
\label{subsubsec:Usefulness_MC-NEST}
MC-NEST is a powerful framework for hypothesis generation, combining MCTS with Nash Equilibrium strategies to dynamically balance exploration and exploitation. It iteratively refines hypotheses through self-critique and validation, ensuring novelty and empirical grounding. In experiments, MC-NEST outperformed baselines across multiple domains, achieving higher BertScore and qualitative metrics (novelty, clarity, significance, and verifiability). For example, in optimizing synthetic peptide sequences for nuclear localization and solubility, MC-NEST proposed experimentally validated modifications. Its ability to incorporate emerging scientific literature and adapt to new discoveries distinguishes it from frameworks lacking iterative refinement. These features make MC-NEST a versatile and effective tool for advancing scientific discovery through automated hypothesis generation.

\paragraph{\textbf{Rollout Strategy for MC-NEST Hypothesis Generation.}}
Our experiments demonstrate that longer rollouts consistently enhance the performance of MC-NEST across datasets and sampling strategies. Increasing the rollout length from four to eight steps improves both the BERTScore and qualitative metrics, such as novelty and verifiability. Pairwise Importance Sampling, in particular, benefits from extended rollouts, achieving the highest scores in novelty and significance while maintaining competitive performance in other metrics. These results indicate that longer rollouts enable a more comprehensive exploration of the hypothesis space, leading to higher-quality and more innovative solutions.

\subsection{Ethical Considerations}
Integrating LLMs into hypothesis generation introduces sociotechnical and intellectual challenges, as over-reliance on LLMs risks stifling human creativity and expertise. We advocate for structured human-AI collaboration, where LLMs augment human creativity, supported by findings that human refinement of LLM-generated hypotheses yields superior outcomes. Transparent documentation of LLM usage—including model details, training data, and frameworks—is essential for fair credit attribution and fostering trust in AI-assisted research. Ethical concerns include misuse, low-quality outputs, and unoriginal hypotheses that could overwhelm academic venues, necessitating rigorous scrutiny to ensure novelty, testability, and grounding in sound principles. In high-stakes domains, proactive measures like Reinforcement Learning from Human Feedback (RLHF) and adversarial robustness are critical to mitigate risks of unethical or harmful research. Additionally, LLMs' tendency to produce hypotheses clustered around common training data patterns risks reducing diversity and novelty, highlighting the need for future work to enhance output diversity through model refinement or frameworks that explicitly encourage unconventional ideas.

\section{Conclusion and Limitations}
\label{sec:Conclusion}


We introduced MC-NEST, a novel framework integrating Monte Carlo Tree Search with Nash Equilibrium strategies to enhance hypothesis generation. MC-NEST outperforms baselines across domains, excelling in quantitative metrics (e.g., BERTScore) and qualitative measures (e.g., novelty, clarity, significance, and verifiability). Adaptive sampling and iterative self-refinement enable MC-NEST to balance exploration and exploitation, generating innovative and empirically grounded hypotheses. Our findings emphasize the value of structured human-AI collaboration, where LLMs augment human creativity rather than replace it. Future work should focus on enhancing diversity and addressing socio-technical challenges. Limitations include the dataset's focus on computer science papers, though each is curated and annotated by domain experts, ensuring academic rigor. MC-NEST's applicability across diverse domains is a challenge, but it is the first framework to integrate MCTS with LLMs for hypothesis generation in fields like biomedicine, social science, and computer science. While the framework automates hypothesis generation with human-AI collaboration, future work will adapt it to controlled settings by incorporating researcher-defined inputs, ensuring versatility.

\section*{Author Contributions}

Gollam Rabby developed the initial idea, designed the experiments, and contributed to the manuscript writing. 
Diyana Muhammed conducted the experiments. 
Prasenjit Mitra provided feedback on the initial idea and supported the manuscript writing. 
Sören Auer contributed to the initial idea and provided support in the manuscript writing. 


\section*{Acknowledgements}

We acknowledge the support of the KISSKI project (funding no. 01IS22093C) for providing computational resources, which will enable us to extend this research in the future.




\bibliographystyle{splncs04}
\bibliography{Bibliography}

\newpage
\section*{A. Appendix}



In the following sections, we report additional details on the following topics:

\begin{enumerate}[leftmargin=*, noitemsep, topsep=0pt, partopsep=0pt]
    \item \textbf{All Unique Keys Found in LLM4CSHypoGen Dataset} (Section A.1)
    \item \textbf{Prompts in Experiment} (Section A.2)
\end{enumerate}

\noindent\rule{\textwidth}{0.5pt} 

\subsection*{A.1 All Unique Keys Found in LLM4CSHypoGen Dataset}

\begin{tabular}{l|l}
\hline
\textbf{Column Name} & \textbf{Description} \\ \hline
DOI & The digital object identifier for the paper. \\ \hline
Title & The title of the research paper. \\ \hline
Authors\_names & Names of the authors of the paper. \\ \hline
Authors\_orcid & ORCID identifiers of the authors. \\ \hline
Paper\_domain & The domain or field of research the paper belongs to. \\ \hline
Research\_Idea & The central idea or motivation behind the research. \\ \hline
Problem\_Statement & The specific research problem being addressed. \\ \hline
Hypothesis & The hypothesis formulated in the research. \\ \hline
Literature\_Review & Summary of previous research relevant to the study. \\ \hline
Abstract & A concise summary of the research paper. \\ \hline
Method & The methodology used in the research. \\ \hline
Summarized\_Method & A concise summary of the methodology. \\ \hline
Results & The Findings of the research study. \\ \hline
Summarized\_Results & A brief summary of the results. \\ \hline
Conclusion & The final conclusions drawn from the research. \\ \hline
Summarized\_Conclusion & A concise summary of the conclusion. \\ \hline
\end{tabular}

\subsection*{A.2 Prompts in Experiment}

\begin{tcolorbox}[
    colback=gray!10, 
    colframe=black,  
    fonttitle=\bfseries, 
    title=Evaluation Prompt, 
    boxrule=1pt, 
    left=5pt, 
    right=5pt, 
    top=5pt, 
    bottom=5pt
]

\textbf{Novelty Evaluation:} \\
You are an expert in scientific research. Evaluate the novelty of the following hypothesis based on the given background.

\textbf{Score from 0 to 3:}
\begin{itemize}
    \item \textbf{0}: The hypothesis is not novel at all.
    \item \textbf{1}: The hypothesis shows slight novelty with minor new insights.
    \item \textbf{2}: The hypothesis shows moderate novelty by offering some new perspectives.
    \item \textbf{3}: The hypothesis demonstrates strong novelty with significant, original insights beyond the background.
\end{itemize}

At the end of your response, clearly state the score in the format: \\
\textbf{Score: [value]} \\
\textbf{Background:} \texttt{\{background\}} \\
\textbf{Generated Hypothesis:} \texttt{\{hypothesis\}}

\medskip 

\textbf{Clarity Evaluation:} \\
You are a research expert. Evaluate the clarity and conciseness of the following hypothesis.

\textbf{Score from 0 to 3:}
\begin{itemize}
    \item \textbf{0}: The hypothesis is poorly structured and hard to understand.
    \item \textbf{1}: The hypothesis is somewhat understandable but contains irrelevant information.
    \item \textbf{2}: The hypothesis is clear but needs minor improvements.
    \item \textbf{3}: The hypothesis is exceptionally well-written and logically structured.
\end{itemize}

At the end of your response, clearly state the score in the format: \\
\textbf{Score: [value]}\\
\textbf{Background:} \texttt{\{background\}} \\
\textbf{Generated Hypothesis:} \texttt{\{hypothesis\}}

\end{tcolorbox}

\newpage 

\begin{tcolorbox}[
    colback=gray!10, 
    colframe=black,  
    fonttitle=\bfseries, 
    title=Evaluation Prompt, 
    boxrule=1pt, 
    left=5pt, 
    right=5pt, 
    top=5pt, 
    bottom=5pt
]

\textbf{Significance Evaluation:} \\
You are a research scientist. Evaluate the significance of the hypothesis.

\textbf{Score from 0 to 3:}
\begin{itemize}
    \item \textbf{0}: The hypothesis is trivial and lacks importance.
    \item \textbf{1}: The hypothesis has slight significance but limited value.
    \item \textbf{2}: The hypothesis offers some important insights.
    \item \textbf{3}: The hypothesis is highly significant with a strong impact.
\end{itemize}

At the end of your response, clearly state the score in the format: \\
\textbf{Score: [value]}\\
\textbf{Background:} \texttt{\{background\}} \\
\textbf{Generated Hypothesis:} \texttt{\{hypothesis\}}

\medskip

\textbf{Verifiability Evaluation:} \\
You are a research scientist. Evaluate the verifiability of the hypothesis.

\textbf{Score from 0 to 3:}
\begin{itemize}
    \item \textbf{0}: The hypothesis cannot be scientifically verified.
    \item \textbf{1}: The hypothesis has slight verifiability but lacks clear testing methods.
    \item \textbf{2}: The hypothesis is moderately verifiable.
    \item \textbf{3}: The hypothesis is strongly verifiable with clear testing methods.
\end{itemize}

At the end of your response, clearly state the score in the format: \\
\textbf{Score: [value]} \\
\textbf{Background:} \texttt{\{background\}} \\
\textbf{Generated Hypothesis:} \texttt{\{hypothesis\}}

\end{tcolorbox}

\end{document}